\documentclass{article}

\PassOptionsToPackage{numbers}{natbib}
\usepackage{amsmath}
\usepackage{graphicx}

\usepackage[preprint]{neurips_2025}

\usepackage[utf8]{inputenc} % allow utf-8 input
\usepackage[T1]{fontenc}    % use 8-bit T1 fonts
\usepackage{hyperref}       % hyperlinks
\usepackage{url}            % simple URL typesetting
\usepackage{booktabs}       % professional-quality tables
\usepackage{amsfonts}       % blackboard math symbols
\usepackage{nicefrac}       % compact symbols for 1/2, etc.
\usepackage{microtype}      % microtypography
\usepackage{xcolor}         % colors
\usepackage{multirow}

\title{ADAM: Autonomous Discovery and Annotation Model using LLMs for Context-Aware Annotations}

\author{%
  Amirreza Rouhi \\
  Department of Electrical and Computer Engineering\\
  Drexel University\\
  Philadelphia, PA 19104 \\
  \texttt{ar3755@drexel.edu} 
  \And
  Solmaz Arezoomandan \\
  Department of Electrical and Computer Engineering\\
  Drexel University\\
  Philadelphia, PA 19104 \\
  \texttt{sa3747@drexel.edu} 
  \And
  Knut Peterson\\
  Department of Electrical and Computer Engineering\\
  Drexel University\\
  Philadelphia, PA 19104 \\
  \texttt{kp3275@drexel.edu} 
  \And
  Joseph T. Woods \\
  Department of Electrical and Computer Engineering\\
  Drexel University\\
  Philadelphia, PA 19104 \\
  \texttt{jw3897@drexel.edu} 
  \And
  David K. Han \\
  Department of Electrical and Computer Engineering\\
  Drexel University\\
  Philadelphia, PA 19104 \\
  \texttt{dkh42@drexel.edu} 
}

\begin{document}

\maketitle

\begin{abstract}

Object detection models typically rely on predefined categories, limiting their ability to identify novel objects in open-world scenarios. To overcome this constraint, we introduce ADAM: Autonomous Discovery and Annotation Model, a training-free, self-refining framework for open-world object labeling. ADAM leverages Large Language Models (LLMs) to generate candidate labels for unknown objects based on contextual information from known entities within a scene. These labels are paired with visual embeddings from CLIP to construct an Embedding-Label Repository (ELR) that enables inference without category supervision. For a newly encountered unknown object, ADAM retrieves visually similar instances from the ELR and applies frequency-based voting and cross-modal reranking to assign a robust label. To further enhance consistency, we introduce a self-refinement loop that re-evaluates repository labels using visual cohesion analysis and kNN-based majority relabeling. Experimental results on the COCO and PASCAL datasets demonstrate that ADAM effectively annotates novel categories using only visual and contextual signals—without requiring any fine-tuning or retraining.

\end{abstract}

\label{sec:intro}
\section{Introduction}
Humans often learn through context, as shown by their early language acquisition skills. If a child encounters a new word, such as "garnet," and hears, "The garnet shimmered brightly in the sunlight," and another time, "The garnet was like ruby in color," their understanding of meaning develops from the context in which that word is being presented. Over time, the child learns a "garnet" is an expensive, red stone—not by direct instruction but through inference \cite{nagy1985learningwords, webb2008effects}. This principle is central to the ADAM methodology: open-world learning from context. ADAM aims to mimic this human capability—the learning of previously unfamiliar entities in an open-world environment. 

Modern object detection models have achieved considerable success in identifying objects from predefined categories in controlled environments~\cite{lin2024generative, cai2018cascade, carion2020end, he2017mask, ren2016faster, sun2021sparse}. However, their utility remains confined to closed-set settings, where all test-time categories are known at training time. These models fundamentally rely on supervised learning over fixed label spaces, rendering them ineffective in open-world environments where novel or long-tail object categories frequently appear. 
%This closed-set assumption limits the flexibility and scalability of systems deployed in dynamic or unfamiliar contexts. 
Recent advances in open-world recognition attempt to address these limitations by leveraging large-scale vision-language models (VLMs)~\cite{radford2021learning, li2021grounded}, prompt-driven classification~\cite{gu2022openvocabulary}, or knowledge-enhanced object detectors~\cite{zang2022open}. While effective, these methods often require costly fine-tuning, handcrafted prompts, or auxiliary annotations to adapt to unseen categories. Moreover, many approaches still assume the presence of training data for base categories or rely on model retraining, restricting real-time adaptability.

To address this challenge, we introduce \textbf{ADAM}: a \textit{training-free}, \textit{self-refining}, and \textit{context-aware} framework for open-world object recognition. ADAM represents a paradigm shift in object detection by moving beyond category-constrained supervision. Instead, it mimics the human-like process of inference through context. ADAM applies this principle to visual recognition. Rather than relying on a fixed label set, ADAM constructs an \textbf{Embedding-Label Repository (ELR)} without any gradient-based updates or task-specific retraining. Starting with region proposals from a pretrained detector, ADAM uses a large language model (LLM) to assign provisional semantic labels based on contextual image captions. These labels and their associated embeddings populate the ELR. Crucially, the system includes a \textit{self-refinement mechanism}: by examining the visual-semantic similarity among embeddings within each proposed class, ADAM revisits noisy predictions and reassigns labels using majority voting over the $k$-nearest neighbors in the embedding space.

\textbf{Our key contributions are as follows:}
\vspace{-0.5em}
\begin{itemize}
    \item We propose a novel zero-shot framework for open-world object labeling that bypasses the need for supervised training on unknown categories.
    \item We design a context-aware prompt construction mechanism that integrates visual descriptors and spatial layouts to effectively guide large language models.
    \item We introduce an embedding-label memory module and a two-stage refinement process combining frequency-based aggregation and cross-modal re-ranking to improve robustness and disambiguate noisy predictions.
    \item We conduct extensive experiments on both COCO and PASCAL VOC datasets, demonstrating that ADAM can accurately annotate novel object categories using only contextual and visual signals, achieving competitive results without any fine-tuning of vision or language backbones.
\end{itemize}
\vspace{-1em}

\section{Related Work}
\label{sec:related}

\subsection{Open Vocabulary and Open World Object Detection}
In recent years, object detection has moved beyond supervised learning, instead focusing on open-set detection. Many current approaches perform Open Vocabulary Detection (OVD), which takes a list of object labels for classification at inference time \cite{Zareian_2021_CVPR}. The limitations of these models such as OWL-ViT, GLIP, Detic, and YOLO-World is that the user must know in advance what objects may appear in the scene \cite{Minderer_2022_Owlvit, li2021grounded, zhou2022detecting, Cheng2024YOLOWorld}. To circumvent the label list, Open World Object Detection (OWOD) methods such as ORE were developed \cite{joseph2021openworld}. A major limitation of these existing OWOD methods is that they require a human oracle to label unknown objects and iteratively retrain the model, which can be costly. 
%Most current approaches to object detection rely on supervised learning and train models to recognize objects from a predefined, closed-set of class labels. Other approaches, such as Open-Vocabulary Detection (OVD) \cite{Zareian_2021_CVPR} are more flexible as their lists of class labels are able to be changed, but are still limited to a provided set of labels or prompts. Open-World Object Detection (OWOD) \cite{joseph2021openworld} differs from this, and is characterized by identifying objects outside its known class list as "unknown", and then progressively adding labels to identify these new unknown objects. 
%One way to achieve this is through incremental learning, enabling the model to progressively recognize and classify new object categories as they emerge. 
%However, previous approaches often relied on an "oracle"---a human in the loop---to aid in novel class labeling \cite{joseph2021openworld}. 
%This approach to object detection was first defined by Joseph et. al.  with the introduction of ORE, a model that used feature clustering and multiple prediction heads for bounding box regression and object classification, as well as an oracle to label machine-proposed unknown objects. 
 
Later works, such as Maaz et. al. \cite{maaz2021classagnostic} and OW-DETR \cite{gupta2021ow} focus on developing class-agnostic object detectors to identify unknown objects, but do not attempt to predict labels for them. OSODD \cite{zheng2022TOWOD} expands work in unknown object detection to cluster previously unseen classes, allowing for novel class grouping and discovery in an unsupervised manner, but they also do not predict labels for the new classes. %Many recent approaches to OVD are also relevant to OWOD tasks, especially in the detection of novel classes. OWL-ViT \cite{Minderer_2022_Owlvit}, GLIP \cite{li2021grounded}, Detic \cite{zhou2022detecting}, and YOLO-World \cite{Cheng2024YOLOWorld} all require a predefined vocabulary or category prompt list during inference, but generic prompts such as "object" can be used for detecting novel classes. 
Many of these works have made great progress in novel class detection and classification from provided sets of labels, however, \textbf{none of these works are capable of predicting novel labels for previously unseen object classes without a class list or oracle}.

% Similarly, GroundingDINO, BLIP, LLaVA , ASMv2 require prompts, referring expressions, or caption-based queries that reference known categories or require instruction tuning.
 % LLaVA-Next, BLIP-3,

\subsection{Vision-Text Features and Contextual Reasoning}

When it comes to predicting novel text labels for new classes, most models are limited to the open-vocabulary approach by using long lists of class labels and matching them to images, such as in CLIP \cite{radford2021clip} or RAM \cite{zhang2023recognize}. Other more recent works have begun exploring ways to surpass this limitation by using LLMs, such as in RAM++ \cite{huang2023open} which uses LLMs to add additional descriptors based on ground truth tag information, expanding the set of tags that can be matched to images. DVDet \cite{jin2024llms} uses a different approach, by matching visual embeddings with fine-grained text descriptors of object parts to improve detections. This approach has also been explored to extract text features of objects, such as texture with CLIP \cite{wu2022doesclipunderstandtexture}, or general descriptors in OvarNet \cite{chen2023ovarnet}. While extracting text features of objects does not directly provide new object class labels, additional textual information can be used to aid in novel label generation.

Other methods combine image and language more effectively into single architectures, such as BLIP \cite{li2022blip} and LLaVa \cite{liu2023llava}, both of which excel at tasks such as visual question answering. These models do not rely on class lists for image captioning, allowing them to generate novel captions of image contents from scratch. However, both models require massive amounts of supervised training data to achieve effective results. Another approach to leveraging LLMs is to use their contextual knowledge to help make predictions where existing labels or visual features are not sufficient. Rouhi et. al. \cite{rouhi2025enhancing} explored this approach using LLaMA \cite{touvron2023llama} to enhance object detection by generating contextually aware labels for occluded or poorly visible objects, achieving significant improvements.

\subsection{Label Refinement and Consistency in Embedding Spaces}
Clustering is a core unsupervised technique for grouping data in high-dimensional spaces without ground truth labels \cite{oyewole2023data, saxena2017review, xie2020hierarchical}, with applications in data organization \cite{cui2021new}, image retrieval \cite{wang2023retccl, anju2022faster}, and anomaly detection \cite{li2021clustering, ariyaluran2022clustering}. Contrastive learning methods like SimCLR \cite{chen2020simple}, SCAN \cite{van2020scan}, and SPICE \cite{niu2022spice} have shown improved clusterability of learned representations. Advanced clustering models such as DeepDPM \cite{ronen2022deepdpm}, ClusterGAN \cite{mukherjee2019clustergan}, DINO-ViT \cite{amir2022deepvitfeaturesdense}, and ClusterFormer \cite{liang2024clusterfomer} leverage pretrained or generative models. However, clustering often struggles with scalability and requires predefined cluster counts \cite{han2022data}, while methods like the elbow criterion are costly at scale \cite{schubert2023stop}.

To overcome these limitations, Nearest Neighbor Search (NNS) is a flexible alternative as it retrieves similar data based on distance metrics without requiring global structure. Works such as \cite{girdhar2023imagebind, amir2022deepvitfeaturesdense} have applied NNS for embedding-based retrieval using cosine similarity. Tools like FAISS \cite{douze2024faiss, johnson2019billion} and SPANN \cite{NEURIPS2021_299dc35e} offer scalable and memory-efficient NNS solutions.

\section{Methodology}
\label{sec:methodology}

Our method of labeling unknown objects in open-world settings leverages contextual information and visual cues to infer object identities without relying on predefined categories. ADAM addresses this challenge through a training-free, self-refining framework that associates unknown object regions with candidate labels based on contextual and visual similarity.

\subsection{Theoretical Motivation}
\label{sec:theoretical}

In open‑world labeling, inferring a semantic label \(Y\) benefits from contextual variables \(X_1,\dots,X_n\) by reducing its conditional entropy:
\[
H(Y \mid X_1,\dots,X_n) \le H(Y \mid X_1,\dots,X_{n-1})\,.
\]
This principle—central to information theory—states that each additional relevant context lowers the uncertainty in \(Y\). It underpins our context‑aware refinement process (see Supplementary Material).

\subsection{System Overview}

\begin{figure*}[t] % The * makes the figure span both columns
\centering
\includegraphics[width=0.9\textwidth]{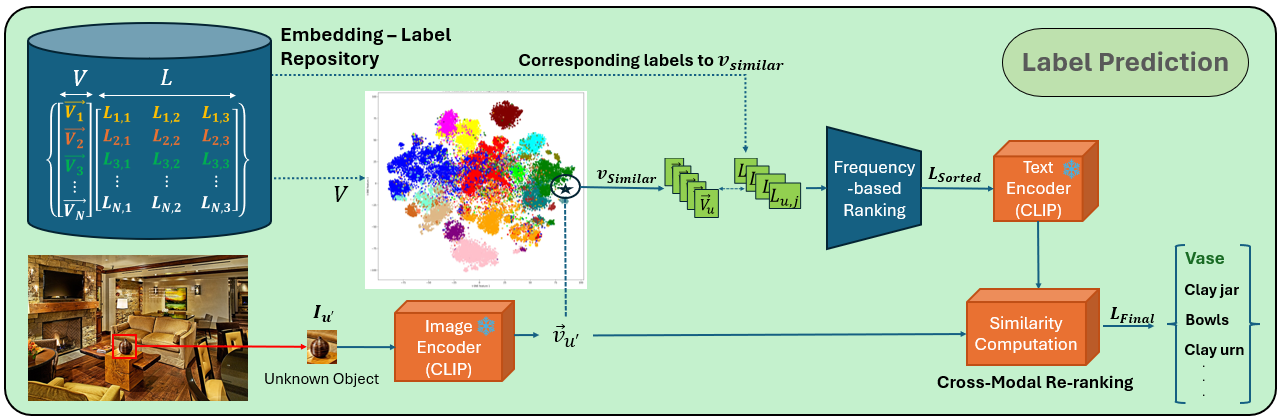}
\caption{\textbf{Schematic of the Label Prediction Process for New Unknown Objects:} The unknown object \( I_{u'} \) is processed through a CLIP image encoder to extract its visual embedding \( \vec{v}_{u'} \). Similar embeddings \( \vec{v}_{\text{similar}} \) are retrieved from the embedding-label repository, and their associated candidate labels are ranked using frequency-based ranking. Finally, CLIP’s text encoder performs cross-modal reranking to refine the labels, yielding the top prediction \( L_{\text{final}} \).}
\label{fig:str2}
\end{figure*}

ADAM is composed of two primary stages: (1) building an Embedding-Label Repository (ELR) and (2) predicting labels for novel unknown objects using similarity-based retrieval and refinement (Figure~\ref{fig:str} and Figure~\ref{fig:str2}). In the first stage, ADAM constructs the ELR by associating visual embeddings of masked unknown regions with candidate labels predicted by a large language model (LLM) prompted with contextual information. In the second stage, when a new unknown object is encountered, its embedding is compared against the repository to retrieve similar entries. The associated label predictions are then refined using frequency-based aggregation and cross-modal reranking.

\subsection{Generating the Embedding-Label Repository (ELR)}
\label{sec: generating_repo}
The embedding-label repository serves as the foundational knowledge base for ADAM, storing visual embeddings of unknown objects and their corresponding predicted labels generated by the LLM. This process is divided into two main components:

\subsubsection{Context-Aware Object Prediction (COP) Module}
The main goal of the COP module is to generate a list of predicted labels for an unknown object. Given an input image, objects are categorized into two groups: a list of detected objects with known labels and a separate list of detected, yet unknown objects. 
%In our current implementation the number of unknown objects per scene is limited to one class, but framework readily supports extension to multiple unknown classes per scene.
%and as we focus on novel label prediction and not object detection, all objects locations and labels are obtained from ground truth annotations, and labels are masked for unknown objects. 

\textbf{Textual Characterization of Object Features.} The cropped image $I_u$ of the unknown object $o_u$ is extracted from the input image $\mathcal{I}$ as follows:
\begin{equation}
I_u = \text{Crop}(\mathcal{I}, \mathcal{B}_u).
\end{equation}
Multiple characteristics are then selected by creating a different caption for each text feature within a given characteristic, and then finding the similarities between embeddings of potential captions and the cropped image embedding using CLIP to select the most relevant features:
\begin{equation}
\mathbf{C}_{u, \text{texture}} = \text{CLIP}_{\text{visual}}^{\text{texture}}(I_u), \mathbf{C}_{u, \text{color}} = \text{CLIP}_{\text{visual}}^{\text{color}}(I_u), \mathbf{C}_{u, \text{material}} = \text{CLIP}_{\text{visual}}^{\text{material}}(I_u).
\end{equation}

 The aggregated description of the unknown object is:
\begin{equation}
\mathbf{C}_u = \{\mathbf{C}_{u, \text{texture}}, \mathbf{C}_{u, \text{color}}, \mathbf{C}_{u, \text{material}}, \dots\}.
\end{equation}
Details of these extractions and a full list of text characteristics (e.g. shape, pattern, ...) are detailed in the supplementary materials.

\textbf{Prompt Construction.} 
To generate a prompt for the LLM, we include the labels and bounding box coordinates of the known objects, represented as $\{L_i, \mathcal{B}_i \ |\ o_i \in \mathcal{O}_{\text{known}}\}$, the aggregated characteristics of the unknown object extracted from the CLIP image encoder, $\mathbf{C}_u$, and the bounding box $\mathcal{B}_u$ of the unknown object. The resulting textual prompt $\mathcal{P}$ is structured as:
\begin{equation}
\mathcal{P} = \text{GeneratePrompt}(\{L_i, \mathcal{B}_i\}, \mathbf{C}_u, \mathcal{B}_u).
\end{equation}
An example of a complete prompt can be seen in Figure \ref{fig:str}.

\textbf{Candidate Label Prediction.} The prompt $\mathcal{P}$ is input to the LLM, which generates a list of candidate labels for the unknown object:
\begin{equation}
L_{u,j} = \text{LLM}(\mathcal{P}),
\end{equation}
where $L_{u,j} = \{\text{label}_1, \text{label}_2, \dots, \text{label}_m\}$ is a set of $m$ candidate labels. For this study, the output of the LLM for each unknown object is a list of 50  potential object labels ($m=50$).

\subsubsection{Visual Embedding Generation with Image Encoder}
The second component of embedding-label repository generation focuses on extracting visual embeddings for the unknown objects using an image encoder. In our framework, we utilize the CLIP image encoder for this purpose.

\textbf{Visual Embedding Computation.} The cropped image $I_u$ of the unknown object is input to the CLIP image encoder to generate its visual embedding:
\begin{equation}
\vec{v}_u = \text{CLIP}(I_u).
\end{equation}
The resulting embedding $\vec{v}_u$ represents the unknown object in a high-dimensional feature space and is used in subsequent similarity searches and label assignments.

\begin{figure*}[t] % The * makes the figure span both columns
\centering
\includegraphics[width=0.9\textwidth]{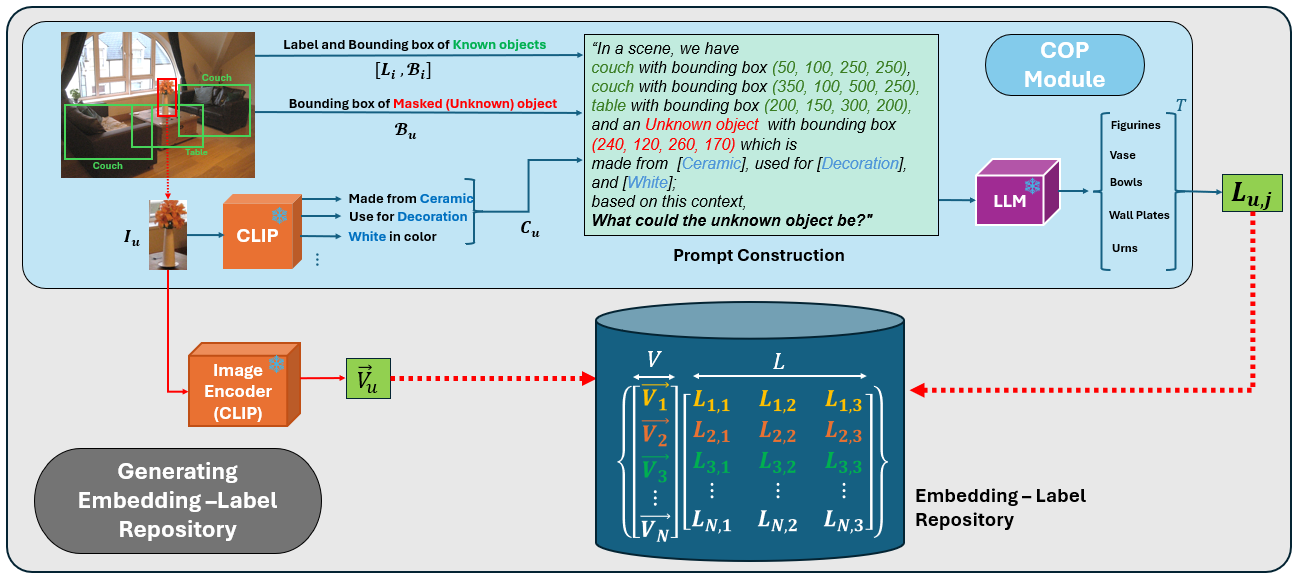}
\caption{\textbf{Generating the Embedding-Label Repository:} This figure illustrates the pipeline for generating the embedding-label repository in ADAM. The Context-Aware Object Prediction (COP) module uses known objects' labels and locations, along with characteristics of the unknown object extracted using CLIP, to construct a context-rich prompt. The prompt is input to an LLM, which generates candidate labels (\(L_{u,j}\)) for the unknown object. Simultaneously, the unknown object's visual embedding (\(\vec{v}_u\)) is computed using the CLIP image encoder. The embedding and corresponding labels are stored in the embedding-label repository for future retrieval and refinement.}

\label{fig:str}
\end{figure*}

The embedding-label repository is maintained as a collection of instance visual embeddings $\{\vec{v}_i \in \mathbb{R}^d\}$, represented as a matrix:
\begin{equation}
V = \begin{bmatrix}
\vec{v}_1 \\
\vec{v}_2 \\
\vdots \\
\vec{v}_N
\end{bmatrix}, \quad V \in \mathbb{R}^{N \times d},
\end{equation}
where $N$ is the total number of objects for which embeddings have been extracted, and $d$ is the dimensionality of each embedding. For each visual embedding, the corresponding LLM-predicted label list is stored in the embedding-label repository as shown in Figure~\ref{fig:str}.

\subsection{Label Prediction for New Unknown Objects}
Assume we have a new unknown object $o_{u'}$ that we want to label. To begin, the object is cropped from the image, and its visual embedding,  $\vec{v}_{u'}$, is calculated based on equations 1 and 8.

\textbf{Similarity Search.} 
We do not directly apply a clustering algorithm, and instead rely on the natural formation of distinct clusters within the CLIP embeddings and use nearest neighbor search to identify similar visual embeddings. Given the query image embedding $\vec{v}_{u'}$, the objective is to identify the top $k$ most similar image indices within the repository. Using FAISS and cosine similarity, this search is defined as:
\begin{equation}
\{ i_1, i_2, \dots, i_k \} = \text{arg top-}k(S(\vec{v}_{u'}, V)),
\end{equation}
where $S$ denotes the cosine similarity function. The indices $\{ i_1, i_2, \dots, i_k \}$ correspond to the $k$ most similar embeddings in $V$.

\textbf{Retrieving Corresponding Labels.} For the query embedding \(\vec{v}_{u'}\), the top \(k\) most similar embeddings are identified from the repository, indexed as \(\{i_1, i_2, \dots, i_k\}\). Each \(\vec{v}_i\) represents the visual embedding of a single object stored in the repository. The corresponding label lists associated with these embeddings are then retrieved as:
\begin{equation}
\mathcal{L}_{\text{similar}} = \{L_{i_1,j}, L_{i_2,j}, \dots, L_{i_k,j}\},
\end{equation}
where \(L_{i,j}\) is the list of candidate labels generated by the LLM for the specific object represented by the embedding \(\vec{v}_i\). %These label lists provide the potential annotations for the unknown object by leveraging its similarity to individual objects within the repository.

\textbf{Frequency-Based Ranking.} The extracted label lists $\mathcal{L}_{\text{similar}}$ are combined and sorted based on the frequency of occurrence of each label across all retrieved lists. Labels that occur in less than 50\% of the label lists are removed. The result is a ranked list of labels $L_{\text{sorted}}$, with the most frequent label appearing first:
\begin{equation}
L_{\text{sorted}} = \text{RankByFrequency}(\mathcal{L}_{\text{similar}}).
\end{equation}

To ensure robustness, we retain only those labels that appear in more than half of the retrieved lists.%, ensuring the robustness of the ranking.
%RERANKING

\textbf{Cross-Modal Reranking} 
We apply CLIP to perform cross-modal reranking \cite{wang-2024-searching} of the candidate labels in $L_{\text{sorted}}$ to improve classification results. For each label $l \in L_{\text{sorted}}$, the CLIP text encoder computes its text embedding:
\begin{equation}
\vec{v}_{\text{text}}(l) = \text{CLIP}_{\text{text}}(l).
\end{equation}
The cross-modal similarity between the previously computed visual embedding $\vec{v}_{u'}$ of the unknown object and each text embedding $\vec{v}_{\text{text}}(l)$ is calculated as:
\begin{equation}
\text{score}(l) = \text{cos}(\vec{v}_{u'}, \vec{v}_{\text{text}}(l)).
\end{equation}
The label with the highest similarity score is then selected as the final prediction:
\begin{equation}
L_{\text{final}} = \arg\max_{l \in L_{\text{sorted}}} \text{score}(l).
\end{equation}
This step ensures that the final label aligns strongly with the visual embedding of the unknown object.

\subsection{Self-Refining Embedding-Label Repository}
To enhance the consistency of label assignments, we extend ADAM with a self-refining mechanism that detects and corrects noisy predictions in the ELR. This process is entirely training-free and operates through unsupervised visual validation and neighborhood-based relabeling.

\textbf{Intra-Class Cohesion Analysis.}
After the initial construction of the repository, we evaluate the visual consistency within each predicted class by computing intra-class cohesion. For each predicted label \( l \), we collect all associated embeddings \( \mathcal{V}_l = \{\vec{v}_i \ |\ l \in L_{i,j}\} \) and calculate the average pairwise cosine similarity:
\begin{equation}
\text{Cohesion}(l) = \frac{1}{|\mathcal{V}_l|^2} \sum_{\vec{v}_i, \vec{v}_j \in \mathcal{V}_l} \cos(\vec{v}_i, \vec{v}_j),
\end{equation}
which quantifies the internal visual agreement among instances of class \( l \).

\textbf{Outlier Detection and Flagging.}
For each embedding \( \vec{v}_u \in \mathcal{V}_l \), we compute its average similarity to all other embeddings in the same class. If this value falls below the class-level cohesion score, the embedding is flagged for potential relabeling:
\begin{equation}
\mathcal{V}_{\text{flagged}} = \{\vec{v}_u \in \mathcal{V}_l \ |\ \text{avg\_sim}(\vec{v}_u) < \text{Cohesion}(l)\}.
\end{equation}

\textbf{Local Relabeling via K-NN Voting.}
Each flagged embedding \( \vec{v}_u \in \mathcal{V}_{\text{flagged}} \) is re-evaluated by querying its \( k_{\text{refinement}} \) nearest neighbors in the embedding space using cosine similarity:
\begin{equation}
\{ i_1, i_2, \dots, i_{k_{\text{refinement}}} \} = \text{arg top-}k(S(\vec{v}_u, V)).
\end{equation}
We aggregate the associated label lists of the retrieved neighbors and apply majority voting to assign a new label:
\begin{equation}
L_{\text{new}} = \text{MajorityVote}(\{L_{i_1,j}, L_{i_2,j}, \dots, L_{i_{k_{\text{refinement}}},j}\}).
\end{equation}
If \( L_{\text{new}} \ne L_{\text{old}} \), we update the label in the repository accordingly.

Throughout all experiments involving repository self-refinement, we set \( k_{\text{refinement}} = 8 \) for the nearest neighbor voting step. This value was empirically selected to balance label stability and correction strength, and further discussion of this choice is provided in the supplementary material.

\textbf{Iterative Refinement.}
This self-refinement process is performed iteratively to ensure semantic consistency across the repository. At each iteration:
\vspace{-0.5em}
\begin{enumerate}
    \item Recompute class-level cohesion scores.
    \item Identify new outliers using updated class assignments.
    \item Relabel flagged instances via \( k_{\text{refinement}} \)-NN majority voting.
\end{enumerate}
\vspace{-0.5em}
The process terminates when either no label changes occur or at a fixed number of iterations \( T \).
%As shown in Section~\ref{sec:ablations}, this iterative refinement improves alignment between semantic labels and visual features.

\section{Experiments}
\label{sec:experiments}

\subsection{Dataset and Evaluation Protocol}
\label{sec:dataset}
To test the ADAM framework, we conducted a comprehensive experimental study using both the COCO 2017 \cite{lin2014microsoft} and PASCAL VOC 2012 \cite{pascal-voc-2012} datasets. 

The \textbf{Emedding-Label Repository} was built using COCO training set. Ground truths were chosen to limit the interference from false positives that occur using an object detector. For an object class $o_u$, we mask its annotations across all images, treating it as the unknown. We then use the remaining set of annotations as the known objects and follow the repository generation process described in Section \ref{sec: generating_repo}. To complete the repository, this process is repeated for every $o_u$ in the dataset and the results are concatenated together.
For generating predicted labels, we utilized the LLaMA v3.2 LLM.

With 860{,}001 annotations in the training set, the Embedding-Label Repository size is \( 860{,}001 \times (50 + 768) \), where each annotation is represented by  \( m = 50 \) LLM-predicted labels and a \( d = 768 \)-dimensional visual embedding. For the \textbf{Label Prediction} step, we set $k = 250$ to retrieve the top $k$ visually similar embeddings during the similarity search. This value was empirically chosen according to section \ref{sec:ablations} (Ablation Study). 

During evaluation, we used the validation sets of COCO and PASCAL VOC. 

It is important to note that our results were obtained without any fine-tuning of the pre-trained CLIP or LLaMA models. For experimentation, we utilized three NVIDIA RTX 3090 GPUs. The average response time from LLaMA for each prompt input was 7.67 seconds.

A key challenge in evaluation is the mismatch between the closed-set evaluation labels and the open-set outputs generated by the LLM, which can include synonyms or semantically similar terms (e.g., "Television" and "TV"). To address this, predicted labels were mapped to ground truth labels using CLIP’s text encoder and cosine similarity. Labels with a similarity of 0.7 or higher to the ground truth were considered correct.

\subsection{Results and Performance Analysis}
The evaluation results of ADAM are presented in Table \ref{tab:mean_accuracy_summary}, which reports the accuracy of the framework for Top-1, Top-3 and Top-5 predictions. Information on accuracy by class is available in the supplemental materials. The mean accuracy for all object categories demonstrates that ADAM achieves promising results in labeling unknown objects, with a mean Top-1 accuracy of 61.30\%. For comparison, we apply either CLIP or BLIP-VQA to the ground truth bounding boxes, providing the list of all COCO labels as input. CLIP significantly underperforms ADAM with a Top-1 accuracy of only 45.65\%. %We also evaluate 
BLIP%-VQA, a strong baseline for visual question answering, which 
achieves the best performance in Top-3 and Top-5, but still falls behind ADAM in Top-1. However, both CLIP and BLIP-VQA require an explicit list of candidate labels or answers at inference time, which limits their applicability in truly open-world scenarios. In contrast, ADAM does not rely on any label list input. Instead, it generates candidate labels dynamically using contextual reasoning and LLM predictions, making it well-suited for environments where novel categories emerge without prior specification. These results validate the effectiveness of the proposed label prediction framework in open-world object detection tasks.

\begin{table}[ht]
\centering
\caption{Mean accuracy on the classification of COCO val boxes. 
Note that ADAM's refinement iterations only apply to its Top-1 accuracy (61.3\%). }
\begin{tabular}{lccc}
\toprule
\textbf{Method} & \textbf{Top-1 Accuracy} & \textbf{Top-3 Accuracy} & \textbf{Top-5 Accuracy} \\
\midrule
CLIP & 45.65 & 64.23 & 71.55 \\
BLIP-VQA & 60.18 & \textbf{92.12} & \textbf{98.16} \\

ADAM   & \textbf{61.30} & 70.64 & 75.95 \\
\bottomrule
\end{tabular}
\label{tab:mean_accuracy_summary}
\end{table}

\subsubsection{Effect of Repository Size on Performance}
\begin{figure}[t]
\centering
\includegraphics[width=0.6\linewidth]{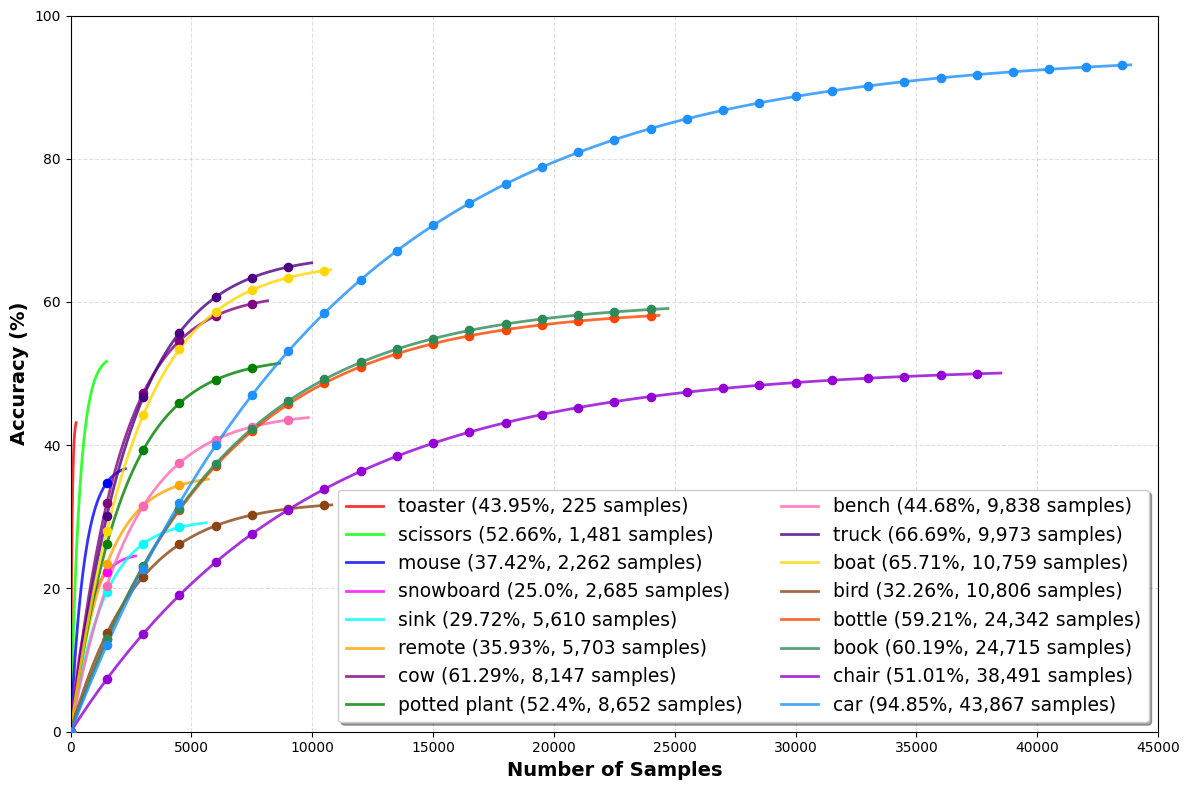} % Replace with your file name
\caption{Top-1 accuracy of ADAM for different object categories as a function of the number of same-class samples in the embedding-label repository \textbf{(Before Self-Refining ELR)}.}
\label{fig:res7}
\end{figure}

Figure~\ref{fig:res7} presents the performance of ADAM as a function of the number of same-class samples in the original embedding-label repository, with each graph line representing a unique object. The results clearly indicate that increasing the number of samples improves the performance of the model. 

This observation aligns with the hypothesis that a larger repository provides more representative embeddings and reduces contextual ambiguity for label prediction. 

This trend underscores the importance of repository size in the proposed framework in open-world object detection tasks.

\subsubsection{Effect of Contextual Information on Accuracy}
The availability of contextual information significantly influences the accuracy of ADAM’s label predictions for unknown objects. Table \ref{tab:accuracy_vs_num_known} illustrates the relationship between the number of known objects in a scene and the model's prediction accuracy. Scenes with no known objects achieve a Top-1 accuracy of only 6\%, as in this case, the method relies solely on the visual characteristics of the unknown object. However, when one to two known objects are present, the accuracy improves to 30\%. As the number of known objects increases beyond eight, the accuracy further rises to 58\%, highlighting the critical role of contextual relationships in enhancing label prediction performance.

\subsection{Evaluating with Region Proposals from Faster R-CNN}
\label{sec:faster}
To evaluate ADAM in a practical detection pipeline and assess its generalization to a new dataset, we apply it to the PASCAL VOC dataset using region proposals from a pretrained Faster R-CNN model \cite{ren2016faster}. The embedding-label repository built from COCO is reused here.
%remains fixed, having been built entirely from COCO training data, making this a true cross-dataset test. 
%For each detected region, we extract its visual embedding and retrieve the top \(k_{\text{detect}} = 8\) nearest neighbors from the repository. The final label is selected via majority voting over their LLM-generated predictions. 
More details on this process 
%and the choice of \(k_{\text{detect}}\) 
are provided in the supplementary material.
For comparison, we apply CLIP and BLIP to the same RPN outputs and provide them with the full COCO label list (which includes all PASCAL categories). In contrast, ADAM operates without any fixed label list.

As shown in Table~\ref{tab:faster}, ADAM outperforms CLIP, yet BLIP has a slight edge in average precision. However, ADAM has better performance than BLIP on most classes, surpassing BLIP in 14 out of 20 classes. This demonstrates ADAM’s strong transferability and its effectiveness as a training-free, open-world labeling solution.

\begin{table}[t]
\centering
\renewcommand{\arraystretch}{1.2}
\setlength{\tabcolsep}{12pt}
\caption{Accuracy of ADAM vs. Number of Known Objects in a Scene \textbf{(Before Self-Refining ELR)}.}

\begin{tabular}{lccccc}
\toprule
\textbf{Number of Known Objects} & \textbf{0} & \textbf{1–2} & \textbf{3–4} & \textbf{5–8} & \textbf{9+} \\
\midrule
\textbf{Average Accuracy (\%)}   & 6.2        & 30.4        & 38.2        & 56.7        & \textbf{58.2} \\
\bottomrule
\end{tabular}
\label{tab:accuracy_vs_num_known}
\end{table}

\begin{table}[tbh]
\centering
\caption{Precision Comparison by Class with FasterRCNN on the PASCAL validation set}
\label{tab:faster}
\large                                  % switched from \small to \medium
\renewcommand{\arraystretch}{1.1}        % more row height
\setlength{\tabcolsep}{3pt}              % tighter column spacing
\resizebox{\linewidth}{!}{%
\begin{tabular}{l*{20}{r}|r}
\toprule
Metric          & person & cat  & dog  & car  & sofa & mbike & train & plane & bus  & horse & bicycle & chair & bird & table & boat & cow  & plant & sheep & tv & bottle & Avg \\
\midrule
FasterRCNN+CLIP   & \textbf{94.8} & 80.0 & 73.1 & \textbf{95.2} & 62.8 & 90.1 & 95.8 & 90.1 & 96.9 & 90.3 & 77.9 & 10.1 & 85.2 & 47.8 & 83.4 & 92.5 & \textbf{93.3} & 86.0 & 17.2 & 47.0 & 75.4 \\
FasterRCNN+BLIP  & 92.3 & 79.5 & 69.9 & 92.8 & \textbf{85.4} & 74.6 & 78.8 & 95.1 & 45.1 & 89.3 & 75.6 & \textbf{74.2} & 88.6 & \textbf{61.6} & 80.0 & 88.8 & 88.6 & \textbf{95.1} & \textbf{71.1} & 60.5 & \textbf{79.3} \\
FasterRCNN+ADAM  & 92.8 & \textbf{82.3} & \textbf{75.1} & 87.5 & 30.2 & \textbf{90.7} & \textbf{99.4} & \textbf{97.7} & \textbf{97.4} & \textbf{91.3} & \textbf{78.1} & 44.6 & \textbf{96.5} & 49.4 & \textbf{94.0} & \textbf{98.5} & 89.2 & 77.0 & 49.3 & \textbf{62.5} & 79.2 \\
\bottomrule
\end{tabular}%
}
\end{table}

\subsection{Ablation Study}
\label{sec:ablations}
We conducted an ablation study, summarized in Table~\ref{tab:ablation_study}, to analyze the contribution of key components in the ADAM framework. The study evaluates the accuracy under various settings across five values of $k$ (0, 50, 150, 250, and 500). Here, $k=0$ represents the baseline where the embedding-label repository is not utilized, effectively disabling the similarity search step.  

\begin{table*}[t]
\centering
\renewcommand{\arraystretch}{1.2}
\caption{
Ablation study for ADAM. The table reports accuracy (\%) for different ablation settings across five $k$ values, where $k=0$ denotes no repository search.
}
\vspace{0.3em}
\resizebox{0.7\textwidth}{!}{%
\begin{tabular}{lccccc}
\toprule
\multirow{2}{*}{Method} & \multicolumn{5}{c}{\textbf{k Values}} \\
\cmidrule(lr){2-6}
 & \textbf{$k=0$} & \textbf{$k=50$} & \textbf{$k=150$} & \textbf{$k=250$} & \textbf{$k=500$} \\
\midrule
\multicolumn{6}{l}{\textit{Prompt Changes}} \\
\midrule
No CLIP Characteristics Descriptor         & 30.13 & 38.20 & 40.62 & 42.14 & 43.48 \\
No Bounding Box for Unknown Object         & 28.45 & 42.88 & 46.32 & 49.91 & 47.76 \\
No Known Object Context                    & 3.13  & 4.47  & 8.07  & 8.24  & 10.80 \\
\midrule
\multicolumn{6}{l}{\textit{Model Structure}} \\
\midrule
No Majority Voting (Top Prediction Only)   & --    & 28.11 & 30.09 & 32.23 & 34.32 \\
No Cross Modal Reranking                   & 29.95 & 32.11 & 43.73 & 45.90 & 45.20 \\
With Embedding-Label Repository (ADAM)     & 33.95 & 45.95 & 53.21 & \textbf{57.64} & 52.98 \\
\textbf{+ Self-Refining (Full Model)}      & \textbf{36.10} & \textbf{47.30} & \textbf{55.30} & \textbf{61.30} & \textbf{54.20} \\
\bottomrule
\end{tabular}
}
\label{tab:ablation_study}
\end{table*}

\textbf{Prompt Changes.} 
Excluding key elements from the COP module prompt results in significant performance degradation. Removing the bounding box of the unknown object reduces accuracy by 7.7\% at $k=250$, emphasizing the importance of spatial information. The absence of CLIP-derived characteristics, such as color and texture, results in a similar drop of over 10\%. The most severe decline occurs when the known object context is excluded, reducing accuracy from 30.13\% to 3.13\% at $k=0$ and from 42.14\% to 8.24\% at $k=250$, highlighting its critical role in contextual reasoning.

\textbf{Model Structure.} 
Disabling majority voting and relying solely on the top prediction leads to a sharp performance decline, achieving only 32.23\% accuracy at $k=250$. This shows the importance of frequency-based ranking in aggregating predicted labels. Removing cross-modal re-ranking also reduces accuracy, particularly for larger $k$ values, with a drop from 53.60\% to 45.90\% at $k=250$, underscoring the need for CLIP-based refinement of predicted labels. With the embedding-label repository, we see significant improvement from no repository search at $k=0$ (33.95\%) to $k=250$ (57.64\%). After $k=250$, performance declines to 52.98\%, suggesting that larger neighborhoods introduce noise from less relevant embeddings. The vast improvement with the addition of the repository demonstrates its critical role in label assignment. Lastly, the self-refinement of the repository improves performance by 3.66\% at $k=250$, highlighting the benefits of the process.

\subsection{Limitations}
\label{sec:limitations}
While ADAM achieves high accuracy across diverse object categories, it faces challenges in scenarios where contextual cues are limited. There is also room to test the method on additional datasets, especially those which have more complete labeling of every object in the scene. Additionally, the use of an LLM for label generation, while improving the set of objects that can be classified, comes at the cost of increased computation.

% \subsection{Comparison to Image Classification}

% \textcolor{red}{TODO: update this section with CLIP comparisons, discussion}

% Since ADAM does not do any bounding box localization, it can be more closely compared to image classification than object detection. Visual attributes and descriptive text have been explored for image classification as an alternative to raw feature-based classification \cite{Yan_2023_ICCV, Maniparambil_2023_ICCV}. While other methods build prompts leveraging the target class label, our prompts are built entirely from visual features extracted by CLIP. Despite this difference, our accuracy of 57.64\% is only slightly below the average accuracy of 59.28\% found in \cite{menon2022visual}. Unfortunately, direct comparison across datasets is not possible due to differences between the classification of cropped images and full images.

% \subsection{Qualitative Analysis}
% \label{sec:qualitative}
% % Embedding plots, before/after label changes, case studies
% \textcolor{violet}{/ I think showing examples is best left to the supplemental material}

% \section{Discussion}
% \label{sec:discussion}
% Impact, generalization, limitations, scalability
%\textcolor{violet}{/ Part of the review's feedback mentioned the cost / scalability using premium LLMs, so I've asked Amir for how many input/outputs tokens we're using to calculate that. If our cost can at least be cheaper than mechanical turk (which as a minimum of \$0.01 per task), I think we'd be in a really good spot to refute that.}
\section{Conclusion}
\label{sec:conclusion}
%\subsection{Conclusion}
ADAM introduces a completely novel approach to open-world object detection that operates in a fully zero-shot setting, utilizing an open-world vocabulary without the need for predefined categories or labeled training data for unknown objects. This framework bridges the gap between vision and language by integrating LLMs with visual embedding methods to infer labels based solely on contextual and visual features.
%The results of our experiments demonstrate the effectiveness of this idea. By leveraging contextual reasoning and diverse visual features, ADAM is capable of labeling unknown objects in diverse scenarios. 
%Importantly, the goal of this work is to establish the viability of the proposed methodology rather than to achieve state-of-the-art performance immediately. 
%We believe that as ADAM is exposed to larger and more diverse datasets, its capabilities as a zero-shot annotation model will improve significantly, laying the foundation for scalable open-world object detection. Additionally, ADAM’s modular design allows for seamless integration with existing object detection or classification models to identify and annotate unclassified region proposals. This integration has the potential to yield substantial performance improvements in traditional object detection pipelines, as it enables these systems to handle previously unknown objects more effectively. 
This study establishes a foundation for future research in open-world object detection, paving the way for more adaptive and intelligent systems that may reduce the burden of human annotation for computer vision datasets.

% % Final summary + future extensions
% ADAM represents a significant step forward in the field of open-world object detection. Its completely zero-shot approach, combined with the use of an open-world vocabulary, showcases the potential for machine learning systems to label unknown objects without predefined categories. While challenges remain, the framework demonstrates that integrating contextual reasoning, visual embeddings, and text feature descriptors for label prediction is an effective strategy for tackling the problem of open-world object detection.
% The primary aim of this work is to validate the effectiveness of the proposed idea. We believe that as ADAM processes more data, it will evolve into a highly robust zero-shot detection model. Furthermore, integrating ADAM with existing object detection models to annotate unclassified region proposals has the potential to significantly enhance the performance of traditional object detection systems. 

% \section*{References}
\bibliography{neurips_2025}
\bibliographystyle{plainnat}

%References follow the acknowledgments in the camera-ready paper. Use unnumbered first-level heading for
%the references. Any choice of citation style is acceptable as long as you are
%consistent. It is permissible to reduce the font size to \verb+small+ (9 point)
%when listing the references.
%Note that the Reference section does not count towards the page limit.
\medskip

{
\small

%[1] Alexander, J.A.\ \& Mozer, M.C.\ (1995) Template-based algorithms for
%connectionist rule extraction. In G.\ Tesauro, D.S.\ Touretzky and T.K.\ Leen
%(eds.), {\it Advances in Neural Information Processing Systems 7},
%pp.\ 609--616. Cambridge, MA: MIT Press.

%[2] Bower, J.M.\ \& Beeman, D.\ (1995) {\it The Book of GENESIS: Exploring
  %Realistic Neural Models with the GEneral NEural SImulation System.}  New York:
%TELOS/Springer--Verlag.

%[3] Hasselmo, M.E., Schnell, E.\ \& Barkai, E.\ (1995) Dynamics of learning and
%recall at excitatory recurrent synapses and cholinergic modulation in rat
%hippocampal region CA3. {\it Journal of Neuroscience} {\bf 15}(7):5249-5262.
%}

%%%%%%%%%%%%%%%%%%%%%%%%%%%%%%%%%%%%%%%%%%%%%%%%%%%%%%%%%%%%
\appendix

\end{document}

##\begin{figure}
  \centering
  \fbox{\rule[-.5cm]{0cm}{4cm} \rule[-.5cm]{4cm}{0cm}}
  \caption{Sample figure caption.}
\end{figure}

\maketitle

% \label{sec:intro}
% \section{Technical Appendices and Supplementary Material}
% Technical appendices with additional results, figures, graphs and proofs may be submitted with the paper submission before the full submission deadline (see above), or as a separate PDF in the ZIP file below before the supplementary material deadline. There is no page limit for the technical appendices.

\section{Theoretical Motivation}
% \textcolor{red}{I'm not sure the best way to cite this here, but this is a source for Shannon-type inequalities: (in comment) }
% \textcolor{violet}{I put the citation as a footnote for now, we can discuss if that is fitting.}
%https://link.springer.com/chapter/10.1007/978-94-017-1043-5_23

% The motivation for our work is based on the concept of conditional entropy applied to scene context. We hypothesize that scene context behaves like an entropic vector of contextual variables \(X_1,\dots,X_n\), where inferring an unknown semantic label \(Y\) benefits from additional contextual variables. Entropic vectors have the submodular property due to the Shannon Inequality - as more information is added, the entropy can only decrease or remain the same:

% \[
% H(Y \mid X_1,\dots,X_n) \le H(Y \mid X_1,\dots,X_{n-1})\,.
% \]

% Our approach for inferring semantic labels using scene context follows from this, as we base our approach on the idea that additional relevant context lowers the uncertainty in \(Y\). We see this idea validated in our results and ablation studies, as the performance of ADAM increases with the number of known objects in the scene. This is also aided by the textual characterization of object features, as object characteristics also add additional context to the scene.
\subsection{Information-Theoretic Perspective on Contextual Learning}

A core theoretical principle behind ADAM is grounded in information theory, particularly the use of conditional entropy to model contextual uncertainty. Let \( Y \) denote the unknown label of an object and \( X_1, X_2, \dots, X_n \) represent the observed contextual variables (e.g., known object labels, spatial locations, or visual attributes) in the scene. The uncertainty in predicting \( Y \) given these variables is captured by the conditional entropy:

\[
H(Y \mid X_1, X_2, \dots, X_n)
\]

From Shannon's Inequality, we know that as more informative context is added, this entropy cannot increase \footcite{shannon_ineq}:

\[
H(Y \mid X_1, X_2, \dots, X_n) \leq H(Y \mid X_1, X_2, \dots, X_{n-1})
\]

This submodular property of entropy implies that the inclusion of each new contextual object narrows the plausible space of labels for \( Y \). In ADAM, this principle is operationalized through prompt engineering: every new known object added to the prompt provides semantic and spatial constraints that help the LLM disambiguate the label for the unknown object. Our empirical ablations confirm this trend—accuracy increases as the number of known objects rises 
% Had to manually put in "2" here since its in another file.
(see Table 2 in the main paper).

\subsection{Visual Semantics and Distributional Hypothesis}

Another motivation stems from the distributional hypothesis in semantics: similar objects tend to occur in similar visual and contextual settings. ADAM leverages this by aligning unknown object embeddings with previously labeled embeddings in the Embedding-Label Repository (ELR). By using cosine similarity in a semantically-aligned embedding space (e.g., CLIP), ADAM approximates the conditional distribution \( P(Y \mid \vec{v}) \) through nearest neighbor retrieval and voting.

Let \( \vec{v}_{u'} \in \mathbb{R}^d \) be the visual embedding of an unknown object. The similarity-based probability of label \( y \) is estimated by:

\[
P(y \mid \vec{v}_{u'}) \approx \frac{1}{Z} \sum_{j \in \text{NN}} w_j \cdot \mathbf{1}(y \in L_j)
\]

where:
\begin{itemize}
    \item \( \text{NN} \) are the $k$-nearest neighbors of \( \vec{v}_{u'} \) in the ELR.
    \item \( w_j = 1 - d(\vec{v}_{u'}, \vec{v}_j) \) is the weight for neighbor $j$ based on cosine distance $d(\cdot,\cdot)$.
    \item \( L_j \) is the set of LLM-predicted labels associated with neighbor $j$.
    \item \( Z \) is the normalizing constant.
\end{itemize}

This defines a smoothed, non-parametric estimation of label likelihoods. It sidesteps the need for a global classifier and instead performs localized reasoning over the ELR. This non-parametric structure also accommodates long-tail categories and label uncertainty.

\subsection{Majority Voting and the Role of \( k \)}

To robustly aggregate labels, we apply a majority vote across the nearest neighbors. From a probabilistic standpoint, let \( p \) be the probability that a randomly retrieved neighbor holds the correct label. The probability that a majority vote yields the correct label among \( k \) neighbors is:

\[
P(\text{majority correct}) = \sum_{i=\lceil k/2 \rceil}^{k} \binom{k}{i} p^i (1 - p)^{k - i}
\]

This binomial tail distribution indicates that increasing $k$ improves robustness, assuming $p > 0.5$. However, large $k$ values risk semantic drift by incorporating less relevant neighbors. Hence, the optimal choice of $k$ balances confidence (through voting consensus) and locality (through similarity in embedding space). In our framework, $k = 250$ for label prediction and $k_{\text{refinement}} = 8$ for refinement are selected based on these trade-offs and confirmed by ablation results.

\subsection{Refinement as Label Denoising}

The self-refinement mechanism in ADAM can be interpreted as an unsupervised label denoising process. Given that the LLM’s initial predictions may be noisy or contextually inconsistent, refinement iteratively aligns labels based on local visual consensus. This mimics Expectation-Maximization (EM)-like behavior where:
\begin{itemize}
    \item The \textbf{E-step} involves estimating whether a label is inconsistent (i.e., below class-level cohesion).
    \item The \textbf{M-step} reassigns a label via majority voting over nearest neighbors.
\end{itemize}

\begin{figure}[h]
    \centering
    \includegraphics[width=0.75\linewidth]{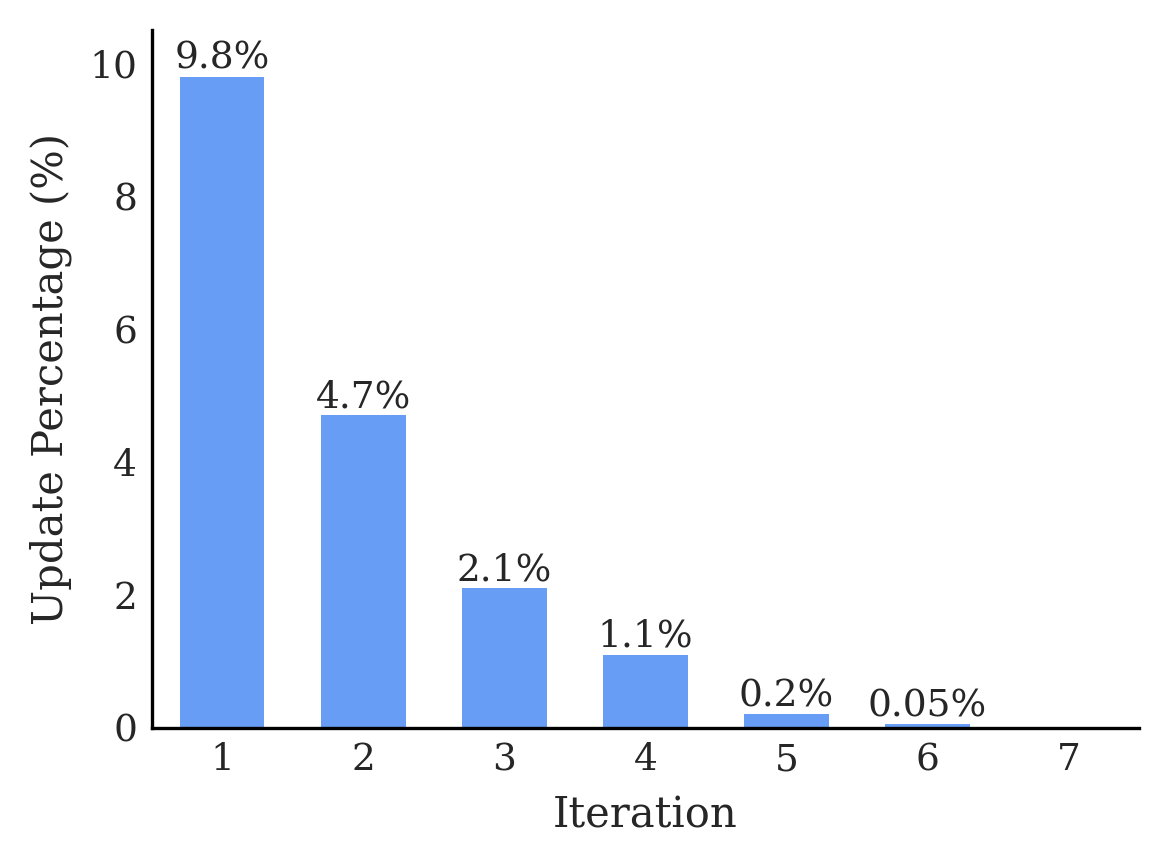}
    \caption{Convergence pattern showing percent of updated labels per iteration. Most updates occur in the first 2–3 iterations.}
    \label{fig:convergence_pattern}
\end{figure}

Through multiple iterations, this refinement reduces entropy in label assignments, increases cohesion within classes, and ultimately improves the repository’s discriminative power. The convergence pattern (Figure~\ref{fig:convergence_pattern}) shows that most updates occur in early iterations, reinforcing that major inconsistencies are correctable with limited passes.

To summarize, ADAM’s theoretical foundation combines:
\begin{itemize}
    \item \textbf{Information theory}: showing that contextual cues reduce label uncertainty via entropy minimization.
    \item \textbf{Non-parametric learning}: using visual embedding similarity for label estimation.
    \item \textbf{Voting theory}: ensuring robustness in the presence of noise via majority aggregation.
    \item \textbf{Unsupervised refinement}: denoising label assignments via iterative consensus.
\end{itemize}

Together, these principles establish the soundness of ADAM’s architecture and explain its strong empirical performance under a training-free paradigm.

\section{Characteristics Used for Object Description}

\subsection{Illustrative Examples}
\begin{figure*}[t!]
\centering
\includegraphics[width=0.8\linewidth]{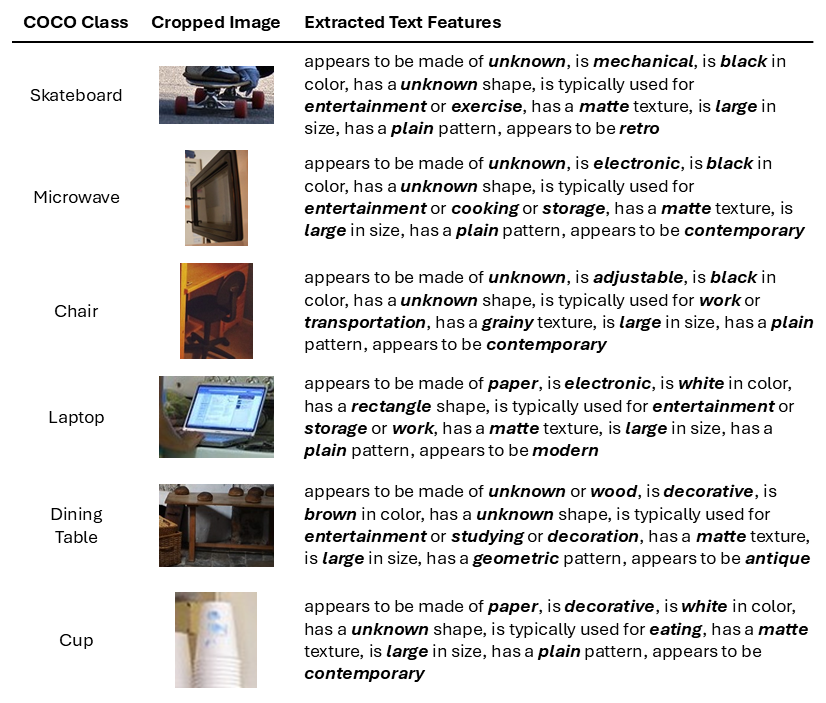}
\caption{Examples of characteristic extraction using CLIP. \emph{Bold} words are features extracted using CLIP. Each object is analyzed for texture, color, material, and other features to generate a detailed description for pseudo-label generation. The extracted text features are then combined into a sentence descriptor to be used in the final LLM prompt.}
\label{fig:examples}
\end{figure*}

Figure~\ref{fig:examples} demonstrates how characteristics are extracted for various objects. For example:
\begin{itemize}
    \setlength\itemsep{-0.4em}
    \item A shiny red car might result in \textit{Shiny} for texture and \textit{Red} for color.
    \item A weathered wooden chair might result in \textit{Weathered} for texture and \textit{Wood} for material.
\end{itemize}
The selected attributes form the textual input for the LLM, providing a rich and contextualized description of the unknown object.

In the Context-Aware Object Prediction (COP) module of ADAM, characteristics of the unknown object play a critical role in generating descriptive prompts for the large language model (LLM). These characteristics are extracted using the CLIP framework, which leverages the visual features of the unknown object and textual descriptions representing specific attributes. Below, we detail how these characteristics are used and provide the categories of attributes considered in our framework.

\subsection{Characteristic Extraction Using CLIP}
For each unknown object, the cropped image $I_u$ of the unknown object is input to the CLIP image encoder to generate its visual embedding:
\begin{equation}
\vec{v}_u = \text{CLIP}(I_u).
\end{equation}

To extract characteristic text features of the object, a list of potential captions are generated, where each caption $c \in L_{\text{captions}}$ is generated using textual features from lists of predefined characteristics. For instance, when extracting the texture of the unknown object, the text features consist of the following list:
\textit{\{Smooth, Rough, Bumpy, Fuzzy, Shiny, Matte, Textured, Polished, Grainy, Glossy, Weathered, Rusted\}.}
For each potential caption $c \in L_{\text{captions}}$, the CLIP text encoder computes its text embedding:
\begin{equation}
\vec{v}_{\text{text}}(c) = \text{CLIP}_{\text{text}}(c).
\end{equation}

The similarity between the previously computed visual embedding $\vec{v}_{u'}$ of the unknown object and each text embedding $\vec{v}_{\text{text}}(c)$ is calculated as:
\begin{equation}
\text{score}(l) = \text{cos}(\vec{v}_{u}, \vec{v}_{\text{text}}(c)).
\end{equation}
The captions in $L_{\text{captions}}$ are then sorted based on their similarity scores, and corresponding text feature of the top caption is selected as the representative attribute for that characteristic:
\begin{equation}
L_{\text{final}} = \arg\max_{c \in L_{\text{captions}}} \text{score}(c).
\end{equation}

This process is repeated for each characteristic category, such as size, color, material, and others. For example:
\begin{itemize}
    \item For the image of a smooth red ball, the texture category's top choice might be \textit{Smooth}, and the color category's top choice might be \textit{Red}.
    \item For an image of a rusted metal pipe, the texture might be \textit{Rusted}, and the material might be \textit{Metal}.
\end{itemize}
Examples of cropped objects and extracted text features can be seen in Figure \ref{fig:examples}.

\subsection{Categories of Characteristics}
The following categories and their respective descriptors were used in the framework:

\subsubsection{Textures}
\textit{\{Smooth, Rough, Bumpy, Fuzzy, Shiny, Matte, Textured, Polished, Grainy, Glossy, Weathered, Rusted\}}

\subsubsection{Sizes}
\textit{\{Tiny, Small, Medium, Large, Huge, Massive\}}

\subsubsection{Patterns}
\textit{\{Solid, Striped, Spotted, Checkered, Camouflage, Plain, Branded, Reflective\}}

\subsubsection{Ages}
\textit{\{Modern, Vintage, Antique, Contemporary, Retro, Weathered\}}

\subsubsection{Functionalities}
\textit{\{Foldable, Adjustable, Portable, Stationary, Electronic, Mechanical, Multi-functional, Decorative, Mobile, Automated, Manual, Emergency, Recreational, Commercial\}}

\subsubsection{Materials}
\textit{\{Metal, Steel, Aluminum, Plastic, Rubber, Glass, Leather, Concrete, Fiberglass, Carbon Fiber, Wood, Fabric, Fur, Feathers, Ceramic, Paper, Cardboard, Foam, Silicone, Unknown\}}

\subsubsection{Colors}
\textit{\{Red, Blue, Green, Yellow, Orange, Purple, Pink, Brown, Black, White, Gray, Cyan, Magenta, Teal, Lavender, Silver, Chrome, Neon, Multi-colored\}}

\subsubsection{Shapes}
\textit{\{Circle, Square, Triangle, Rectangle, Oval, Cylindrical, Aerodynamic, Spherical, Octagonal, Conical, Star, Heart, Crescent, Diamond, Trapezoid, Parallelogram, Rhombus, Unknown\}}

\subsubsection{Use Cases}
\textit{\{Rest, Work, Entertainment, Storage, Cooking, Cooling, Eating, Cleaning, Decoration, Transportation (Public/Private), Communication, Exercise, Studying, Personal Care, Safety, Signaling, Warning, Parking, Sporting, Racing, Emergency, Commercial, Passenger, Cargo\}}

\subsubsection{Environments}
\textit{\{Urban, Rural, Marine, Aerial, Terrestrial, All-terrain, Road, Rail, Water\}}

\subsubsection{Categories}
\textit{\{Vehicle, Animal, Urban Infrastructure, Warning Infrastructure, Vegetation, Sports Equipment, Weather Protection\}}

By systematically extracting and selecting the most likely attribute for each characteristic category, the COP module builds a detailed textual description of the unknown object. This approach ensures that the LLM receives high-quality contextual input, enabling accurate and robust pseudo-label predictions.

\section{Performance Evaluation }
\subsection{Refinement Optimization}
\label{sec:refinement_optimization}
The refinement begins by grouping embeddings by their assigned labels. For each group, intra-class cohesion is computed as the average pairwise cosine similarity among embeddings. Embeddings falling below the cohesion threshold are flagged as inconsistent and re-evaluated using their $k_{\text{refinement}}$ nearest neighbors in the embedding space. Weighted majority voting over the neighbors’ label sets yields a new candidate label for each outlier. Labels are updated only if the candidate differs from the original. The full pseudocode is shown below:

\vspace{0.5em}
\noindent

\begin{algorithm}
\caption{ELR Self Refinement}
\begin{algorithmic}[1]
\State Initialize iteration counter $t = 0$
\Repeat
    \State Set \texttt{update\_count} = 0
    \For{each unique label $l$ in the repository}
        \State $V_l = \{ \vec{v}_i \mid l \in L_{i,j} \}$
        \State Compute cohesion: $\texttt{cohesion}(l)$
        \For{each $\vec{v}_u$ in $V_l$}
            \State Compute avg similarity: $\texttt{avg\_sim}(\vec{v}_u)$
            \If{$\texttt{avg\_sim}(\vec{v}_u) < \texttt{cohesion}(l)$}
                \State Add $\vec{v}_u$ to $V_\texttt{flagged}$
            \EndIf
        \EndFor
    \EndFor
    \For{each $\vec{v}_u$ in $V_\texttt{flagged}$}
        \State Find $k_{\text{refinement}}$ nearest neighbors: $\texttt{NN} = \{i_1, ..., i_k\}$
        \State Retrieve label sets $L_{\texttt{NN}}$
        \State Compute $L_{\text{new}}$ via weighted voting
        \If{$L_{\text{new}} \ne$ current label of $\vec{v}_u$}
            \State Update label of $\vec{v}_u$
            \State Increment \texttt{update\_count}
        \EndIf
    \EndFor
    \State $t \gets t + 1$
\Until{\texttt{update\_count} = 0 or $t \ge T$}
\end{algorithmic}
\end{algorithm}

\textbf{Weighted Majority Voting.} We adopt a similarity-weighted voting scheme where neighbors closer in the embedding space contribute more to the vote. The weight for a neighbor embedding $\vec{v}_j$ with respect to the flagged embedding $\vec{v}_u$ is:
\[
w_j = 1 - d(\vec{v}_u, \vec{v}_j)
\]
where \(d(\cdot, \cdot)\) is the cosine distance between visual embeddings $\vec{v}_u$ and $\vec{v}_j$. The vote for candidate label \(c\) is:
\[
\text{vote}(c) = \sum_{j \in \text{NN}} w_j \cdot \mathbf{1}(c \in L_j)
\]
where $L_j$ is the label set of the $j^{th}$ neighbor, and $\mathbf{1}(c \in L_j)$ is an indicator function that is 1 if $c$ appears in $L_j$, and 0 otherwise.

\textbf{Convergence Criteria.} Refinement stops when no label updates occur during an iteration or when a maximum number of iterations \(T\) is reached. The convergence condition is:
\[
\sum_{i=1}^{|V|} \mathbf{1}(L_{i,j}^{(t)} \ne L_{i,j}^{(t-1)}) = 0
\]
where $|V|$ is the total number of embeddings, and $L_{i,j}^{(t)}$ denotes the label for embedding $i$ at iteration $t$. The indicator is 1 if the label changed from the previous iteration and 0 otherwise.  In our experiments, we set the maximum number of iterations to T=10, which allows sufficient opportunity for correction without excessive computation. 

\begin{figure}[h]
    \centering
    \includegraphics[width=0.95\linewidth]{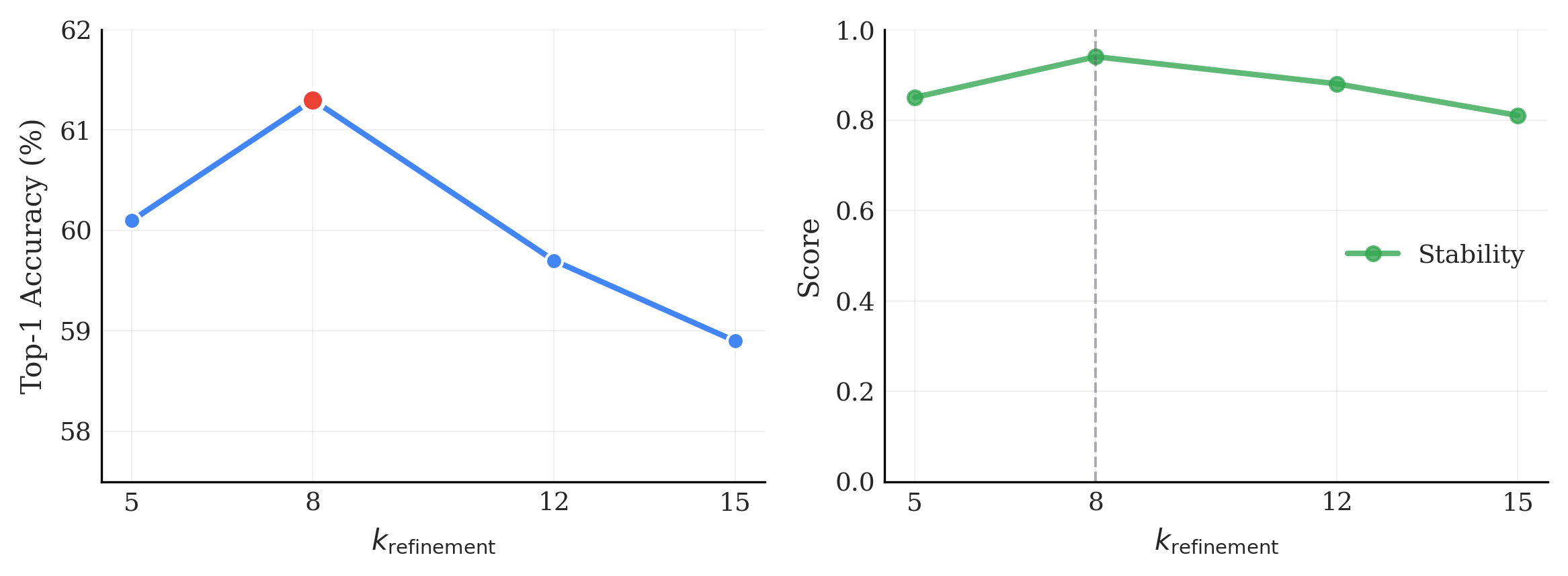}
    \caption{(Left) Accuracy vs. $k_{\text{refinement}}$, showing a peak at $k = 8$. (Right) Stability score across $k$, with $k = 8$ achieving highest consistency.}
    \label{fig:param_analysis}
\end{figure}

\textbf{Hyperparameter Selection.} We conducted a detailed sensitivity analysis (Figure~\ref{fig:param_analysis}) to identify the optimal setting for $k_{\text{refinement}}$. We observed that $k = 8$ strikes a balance between update stability and correction effectiveness. Smaller values lead to unstable noisy updates, while higher values risk oversmoothing and missing corrections.

\textbf{Stability Analysis.} To assess label consistency over iterations, we define a stability metric:
\[
\text{Stability}(t) = \frac{1}{|V|} \sum_{i=1}^{|V|} \mathbf{1}(L_{i,j}^{(t)} = L_{i,j}^{(t-1)})
\]
where $L_{i,j}^{(t)}$ denotes the label of the $i^{th}$ embedding at iteration $t$, and $|V|$ is the total number of embeddings. This measures the proportion of embeddings that retain their label between two consecutive refinement iterations. A value of 1.0 indicates full stability, while lower values reflect volatility. As shown in Figure~\ref{fig:param_analysis}, stability is maximized when $k_{\text{refinement}} = 8$, balancing label preservation with effective correction. We find that 94\% of instances maintain consistent labels after each iteration at this setting.

\begin{figure}[h]
    \centering
    \includegraphics[width=0.75\linewidth]{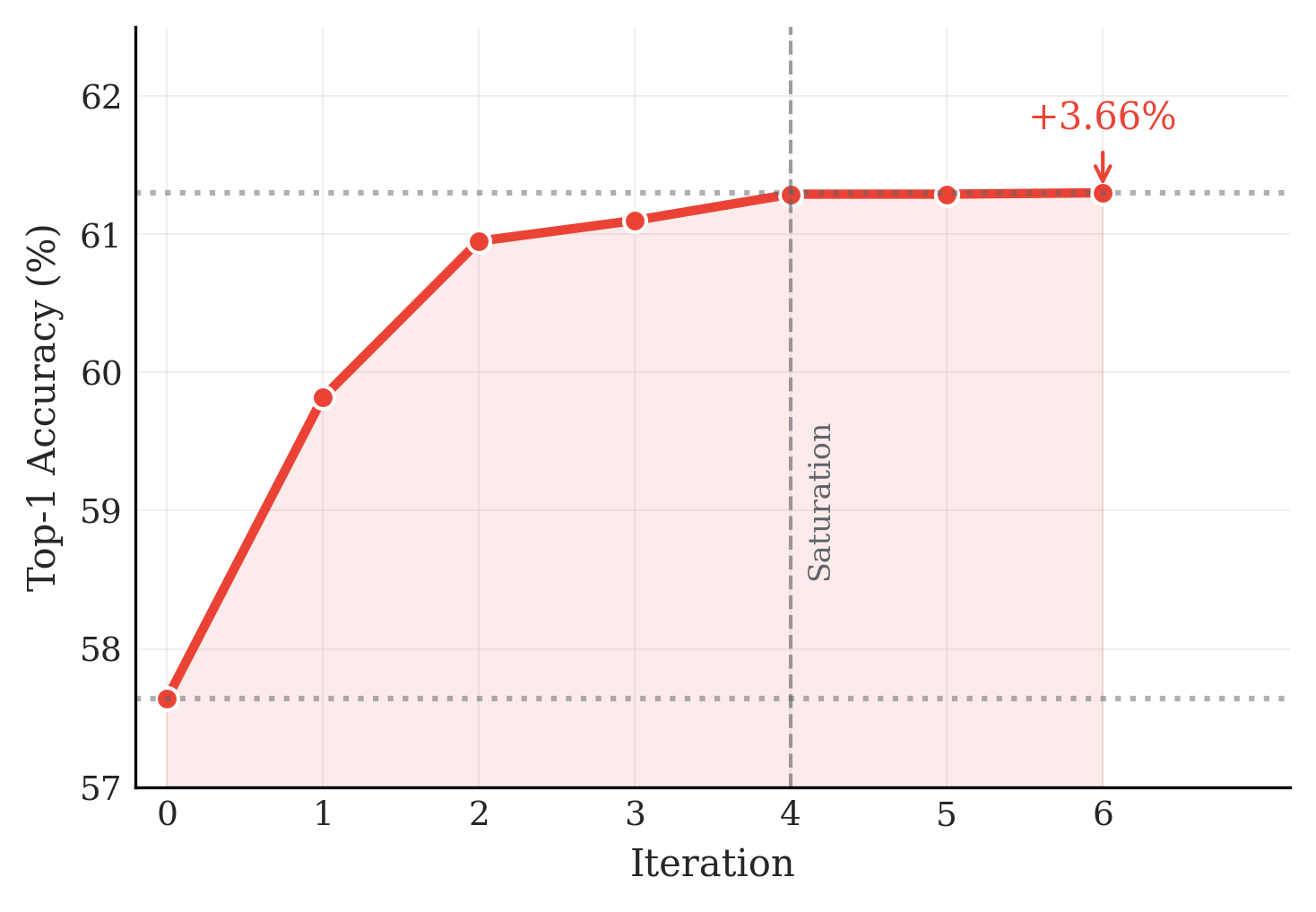}
    \caption{Effect of iterative refinement on Top-1 accuracy. ADAM improves by 3.66\%, with diminishing gains after 4 iterations.}
    \label{fig:self_ref_res}
\end{figure}

\textbf{Empirical Impact.} Figure~\ref{fig:self_ref_res} shows the improvement in Top-1 accuracy of ADAM due to self-refinement, increasing from 57.64\% to 61.30\%. Accuracy saturates after 5–6 iterations, aligning with our convergence criteria and update threshold.

\subsection{By-category results on COCO}
Table \ref{tab:full_results} provides a comprehensive summary of the performance of ADAM, CLIP, and BLIP on unseen objects across all categories in the COCO dataset. CLIP was prompted simply with the list of 80 COCO classes, and a softmax of the CLIP outputs was used to determine the final prediction. For BLIP, the question was asked for each class, "Is this a \{class\}?" If the model replied "yes" then it was given a score of 100\%, if the class appeared in the answer it was given a score of 80\%, and otherwise it was given a score of 0\%. 

ADAM performs well across most categories, achieving high Top-1 accuracy for objects such as `car` (95.11\%), `stop sign` (85.21\%), and `bicycle` (69.16\%). However, the method exhibits comparatively lower performance for categories like `zebra` and `giraffe`, where Top-1 accuracy is 0.00\%. The poor performance for these categories stems from their images lacking other detected objects in the scene. For images containing zebra and giraffe, 74.80\% and 67.67\% are single-object scenes, respectively. In contrast, categories like `car` benefit from co-occurring objects providing the context, enabling effective label prediction. 
CLIP and BLIP have better performance on these objects since they do not rely on context to generate their answers. However, they are also constrained by a label list and still struggle with many classes where ADAM succeeds, such as "spoon", ultimately leading to ADAM outperforming both other methdods overall.

\begin{table*}[ht]
\centering
\caption{Performance of ADAM, ADAM with refinement (ADAM-R), CLIP, and BLIP on unseen object categories. The table reports the accuracy (\%) for Top-1, Top-3, and Top-5 predictions for each object category, sorted alphabetically. \\}
\resizebox{0.9\textwidth}{!}{%
\begin{tabular}{lcccc ccc ccc}
\toprule
\textbf{Category} & \textbf{ADAM-R Top-1} & \textbf{ADAM Top-1} & \textbf{ADAM Top-3} & \textbf{ADAM Top-5} & \textbf{CLIP Top-1} & \textbf{CLIP Top-3} & \textbf{CLIP Top-5} & \textbf{BLIP Top-1} & \textbf{BLIP Top-3} & \textbf{BLIP Top-5} \\
\midrule
airplane & 91.14 & 76.35 & 81.19 & 93.04 & 84.51 & 90.85 & 92.96 & 85.77 & 95.70 & 98.83 \\
apple & 51.29 & 51.29 & 69.78 & 76.34 & 55.88 & 77.94 & 83.82 & 70.83 & 94.77 & 98.83 \\
backpack & 67.15 & 67.15 & 78.34 & 85.12 & 17.31 & 31.94 & 40.30 & 46.39 & 88.85 & 97.94 \\
banana & 83.46 & 83.46 & 90.56 & 94.12 & 66.04 & 79.41 & 85.83 & 70.95 & 93.91 & 98.56 \\
baseball bat & 57.30 & 57.30 & 70.56 & 76.98 & 60.77 & 83.85 & 91.54 & 63.62 & 93.49 & 98.07 \\
baseball glove & 60.16 & 60.16 & 75.42 & 80.87 & 61.72 & 78.91 & 86.72 & 30.53 & 82.25 & 95.74 \\
bear & 50.68 & 50.68 & 67.23 & 72.85 & 83.10 & 90.14 & 92.96 & 83.20 & 95.70 & 98.83 \\
bed & 62.51 & 56.66 & 72.10 & 78.30 & 53.99 & 74.85 & 78.53 & 55.20 & 95.12 & 98.83 \\
bench & 44.68 & 44.68 & 59.57 & 65.25 & 28.30 & 45.05 & 52.47 & 46.19 & 84.40 & 95.02 \\
bicycle & 69.16 & 63.72 & 83.90 & 86.85 & 46.71 & 55.92 & 58.55 & 62.78 & 91.92 & 97.52 \\
bird & 35.48 & 32.26 & 41.94 & 51.61 & 57.33 & 66.00 & 67.67 & 68.16 & 89.32 & 97.18 \\
boat & 83.54 & 65.71 & 81.43 & 82.86 & 38.12 & 46.69 & 50.83 & 61.51 & 86.45 & 96.92 \\
book & 58.91 & 60.19 & 75.85 & 87.65 & 22.12 & 28.07 & 31.48 & 39.41 & 81.19 & 95.83 \\
bottle & 64.51 & 59.21 & 72.34 & 78.62 & 26.01 & 35.38 & 39.31 & 50.43 & 87.52 & 97.92 \\
bowl & 47.02 & 33.25 & 49.86 & 56.72 & 33.05 & 52.54 & 58.64 & 55.92 & 90.84 & 98.32 \\
broccoli & 62.71 & 62.71 & 76.81 & 83.45 & 74.20 & 92.04 & 94.59 & 81.04 & 93.88 & 98.51 \\
bus & 59.87 & 50.31 & 75.62 & 81.48 & 78.57 & 88.57 & 90.36 & 81.15 & 95.70 & 98.83 \\
cake & 55.52 & 55.52 & 70.23 & 77.18 & 56.91 & 69.45 & 73.95 & 66.04 & 91.09 & 97.55 \\
car & 95.11 & 94.85 & 97.82 & 98.15 & 46.73 & 61.67 & 66.54 & 56.77 & 90.56 & 97.54 \\
carrot & 62.27 & 62.27 & 77.23 & 83.69 & 53.87 & 81.66 & 88.25 & 47.66 & 89.67 & 97.41 \\
cat & 76.68 & 76.68 & 85.12 & 90.34 & 82.67 & 94.06 & 96.04 & 86.83 & 95.70 & 98.83 \\
cell phone & 64.10 & 56.50 & 72.06 & 77.83 & 33.18 & 47.73 & 52.73 & 68.57 & 95.27 & 98.83 \\
chair & 51.01 & 51.01 & 66.23 & 71.89 & 21.04 & 36.31 & 41.32 & 48.04 & 89.18 & 97.97 \\
clock & 83.15 & 79.65 & 84.95 & 86.70 & 60.25 & 67.21 & 71.31 & 75.24 & 92.56 & 98.02 \\
couch & 55.99 & 49.30 & 62.80 & 68.39 & 45.59 & 67.82 & 72.41 & 48.40 & 93.50 & 98.83 \\
cow & 64.51 & 61.29 & 70.97 & 77.42 & 66.18 & 78.32 & 82.95 & 66.19 & 92.66 & 97.97 \\
cup & 64.51 & 58.89 & 75.12 & 81.36 & 25.36 & 43.25 & 51.18 & 59.87 & 93.32 & 98.71 \\
dining table & 35.11 & 26.26 & 38.65 & 45.35 & 35.70 & 60.82 & 71.41 & 39.73 & 92.37 & 97.93 \\
dog & 52.59 & 52.59 & 69.12 & 74.81 & 72.90 & 91.59 & 93.46 & 84.08 & 95.70 & 98.83 \\
donut & 63.69 & 63.69 & 78.65 & 85.12 & 41.07 & 51.79 & 59.52 & 60.85 & 92.85 & 98.53 \\
elephant & 85.71 & 53.57 & 53.57 & 57.14 & 80.39 & 83.92 & 86.67 & 86.96 & 95.33 & 98.83 \\
fire hydrant & 71.88 & 71.88 & 82.81 & 87.50 & 75.82 & 82.42 & 83.52 & 82.89 & 95.70 & 98.83 \\
fork & 54.55 & 54.55 & 68.76 & 75.23 & 23.41 & 44.88 & 59.02 & 52.76 & 91.03 & 97.86 \\
frisbee & 50.00 & 50.00 & 50.00 & 50.00 & 50.98 & 62.75 & 69.61 & 66.83 & 92.89 & 98.83 \\
giraffe & 0.00 & 0.00 & 2.94 & 5.09 & 89.96 & 93.01 & 94.32 & 88.50 & 95.70 & 98.83 \\
hair drier & 44.93 & 44.93 & 61.14 & 68.45 & 36.36 & 54.55 & 54.55 & 57.84 & 95.70 & 98.83 \\
handbag & 56.83 & 56.83 & 71.47 & 79.13 & 25.21 & 44.28 & 52.75 & 36.39 & 90.23 & 97.99 \\
horse & 81.30 & 75.51 & 81.63 & 81.63 & 73.61 & 85.13 & 85.87 & 76.70 & 94.63 & 98.46 \\
hot dog & 51.28 & 51.28 & 68.34 & 73.45 & 47.93 & 68.60 & 76.03 & 25.54 & 89.38 & 98.83 \\
keyboard & 53.25 & 50.30 & 65.87 & 73.57 & 66.67 & 78.67 & 87.33 & 55.74 & 91.88 & 97.51 \\
kite & 42.86 & 42.86 & 57.14 & 57.14 & 45.20 & 60.80 & 69.60 & 62.89 & 91.88 & 98.83 \\
knife & 61.29 & 61.29 & 74.58 & 80.23 & 22.73 & 60.74 & 80.17 & 34.55 & 87.40 & 95.15 \\
laptop & 77.66 & 77.66 & 84.11 & 88.10 & 69.70 & 80.09 & 85.28 & 70.03 & 93.64 & 98.83 \\
microwave & 53.65 & 51.68 & 65.65 & 72.54 & 40.00 & 61.82 & 65.45 & 74.36 & 93.96 & 98.83 \\
motorcycle & 71.77 & 68.13 & 83.70 & 88.56 & 73.33 & 81.39 & 85.56 & 82.30 & 95.17 & 98.83 \\
mouse & 52.51 & 37.42 & 48.31 & 53.02 & 10.64 & 14.89 & 21.28 & 47.38 & 90.61 & 98.83 \\
orange & 57.48 & 57.48 & 72.89 & 79.11 & 71.98 & 82.10 & 86.77 & 70.72 & 93.10 & 98.44 \\
oven & 63.41 & 56.32 & 75.05 & 80.73 & 44.37 & 59.15 & 71.13 & 64.00 & 95.70 & 98.83 \\
parking meter & 47.06 & 47.06 & 61.76 & 69.12 & 54.55 & 67.27 & 78.18 & 74.36 & 92.22 & 98.83 \\
person & 78.85 & 78.85 & 85.23 & 90.14 & 45.38 & 73.18 & 84.72 & 61.40 & 94.78 & 98.71 \\
pizza & 50.31 & 50.31 & 67.92 & 73.23 & 75.54 & 83.09 & 85.25 & 81.40 & 94.67 & 98.83 \\
potted plant & 56.51 & 52.40 & 70.19 & 77.64 & 61.75 & 78.31 & 81.33 & 53.10 & 91.66 & 98.23 \\
refrigerator & 61.13 & 56.82 & 70.56 & 75.68 & 65.87 & 76.98 & 80.16 & 75.01 & 95.70 & 98.83 \\
remote & 47.51 & 35.93 & 52.81 & 61.34 & 12.10 & 20.56 & 24.19 & 46.18 & 86.06 & 96.83 \\
sandwich & 68.84 & 68.84 & 80.72 & 88.31 & 64.41 & 85.88 & 88.70 & 75.48 & 93.54 & 98.83 \\
scissors & 52.66 & 52.66 & 67.58 & 74.15 & 38.89 & 61.11 & 63.89 & 70.69 & 95.70 & 98.83 \\
sheep & 86.66 & 84.44 & 84.44 & 86.67 & 63.51 & 76.15 & 80.75 & 70.25 & 91.86 & 97.69 \\
sink & 48.55 & 29.72 & 47.74 & 54.02 & 29.52 & 41.90 & 44.29 & 59.72 & 92.06 & 97.41 \\
skateboard & 30.70 & 30.70 & 55.43 & 60.24 & 43.37 & 57.23 & 65.06 & 68.43 & 89.36 & 97.64 \\
skis & 64.70 & 64.70 & 78.23 & 83.92 & 49.39 & 80.49 & 87.20 & 48.77 & 86.95 & 95.21 \\
snowboard & 25.00 & 25.00 & 50.00 & 50.00 & 33.96 & 58.49 & 69.81 & 56.58 & 90.29 & 98.83 \\
spoon & 77.69 & 77.69 & 88.12 & 92.34 & 17.33 & 35.56 & 49.33 & 34.34 & 88.47 & 98.83 \\
sports ball & 56.50 & 56.50 & 74.23 & 79.81 & 43.21 & 67.28 & 72.84 & 39.83 & 77.98 & 95.17 \\
stop sign & 85.21 & 84.48 & 89.66 & 91.38 & 71.64 & 77.61 & 77.61 & 85.46 & 95.70 & 98.83 \\
suitcase & 78.67 & 78.67 & 85.67 & 89.14 & 39.06 & 53.20 & 58.25 & 63.64 & 86.36 & 98.16 \\
surfboard & 89.47 & 89.47 & 89.47 & 89.47 & 41.56 & 61.90 & 68.83 & 53.50 & 84.93 & 95.83 \\
teddy bear & 77.66 & 77.66 & 87.82 & 92.41 & 73.30 & 80.63 & 80.63 & 17.60 & 93.70 & 98.31 \\
tennis racket & 76.68 & 76.68 & 84.92 & 88.76 & 66.51 & 86.24 & 93.12 & 74.62 & 94.38 & 98.37 \\
tie & 74.53 & 74.53 & 84.71 & 89.62 & 50.54 & 63.98 & 67.74 & 51.79 & 88.50 & 97.76 \\
toaster & 50.61 & 43.95 & 62.78 & 72.20 & 11.11 & 55.56 & 66.67 & 80.78 & 95.70 & 98.83 \\
toilet & 78.92 & 61.14 & 80.17 & 86.19 & 67.60 & 76.54 & 78.77 & 78.69 & 95.17 & 98.83 \\
toothbrush & 83.50 & 83.50 & 91.26 & 95.43 & 36.73 & 46.94 & 55.10 & 61.21 & 89.85 & 98.83 \\
traffic light & 60.56 & 31.28 & 45.96 & 53.84 & 38.85 & 50.39 & 56.96 & 65.83 & 95.45 & 98.83 \\
train & 67.86 & 67.86 & 89.29 & 96.43 & 84.74 & 93.16 & 95.26 & 84.67 & 95.70 & 98.83 \\
truck & 65.10 & 66.69 & 79.34 & 85.43 & 46.44 & 72.24 & 76.66 & 62.53 & 92.65 & 97.37 \\
tv & 78.51 & 77.35 & 89.54 & 93.16 & 36.75 & 58.66 & 70.67 & 68.08 & 95.02 & 98.83 \\
umbrella & 60.75 & 57.81 & 73.42 & 76.79 & 41.71 & 52.07 & 56.48 & 72.75 & 93.47 & 98.83 \\
vase & 58.75 & 43.00 & 65.59 & 75.35 & 27.13 & 50.00 & 59.30 & 52.13 & 85.32 & 97.29 \\
wine glass & 60.31 & 60.31 & 75.45 & 81.71 & 37.81 & 48.44 & 57.81 & 63.34 & 86.43 & 96.66 \\
zebra & 0.00 & 0.00 & 3.55 & 5.24 & 89.18 & 91.42 & 92.54 & 87.49 & 94.63 & 98.83 \\
\bottomrule
\end{tabular}
}

% \vspace{\baselineskip}

% \resizebox{0.3\textwidth}{!}{%
% \begin{tabular}{lccc}
% \toprule
% \textbf{Summary Metrics} & \textbf{Top-1} & \textbf{Top-3} & \textbf{Top-5} \\
% \midrule
% \textbf{Mean Accuracy}   & 57.64          & 70.64          & 75.95          \\
% \bottomrule
% \end{tabular}
% }
\label{tab:full_results}
\end{table*}

% \subsection{Evaluation with Faster R-CNN}

% \textcolor{violet}{TODO: Add more information on the selection of \(k_{detect}\). Add BLIP to the bar chart and update discussion of the 3 methods to include BLIP.} Figure \ref{fig:faster_prec} shows the precision of ADAM compared to CLIP when evaluating on PASCAL VOC using Faster RCNN for bounding boxes, rather than ground truth. For each detected region, the crop's visual embedding is extracted and the top \(k_{\text{detect}} = 9\) nearest neighbors from the repository are retrieved. The final label is selected via majority voting over the LLM predictions generated based on each visual embedding. 
% Despiste, ADAM's repository is built from the ground truths of COCO, the performance when evaluating on region proposals is still quite strong. This shows that ADAM's inference process is still effective on imprecise prediction boxes, in addition to ground truths. We also see that ADAM performs best overall and has fewer low-performance classes than the other methods.

\section{Selecting $k_{\text{detect}}$ for Region Proposal Evaluation}
\label{sec:k_detect_selection}

To optimize the label assignment accuracy for region proposals, we conducted a sensitivity study on the number of retrieved neighbors $k_{\text{detect}}$ used during nearest neighbor search for ADAM’s label inference. This analysis was performed after applying the self-refinement process to the Embedding-Label Repository (ELR), ensuring maximal consistency and semantic coherence of repository labels.

\begin{figure}[h]
    \centering
    \includegraphics[width=0.75\linewidth]{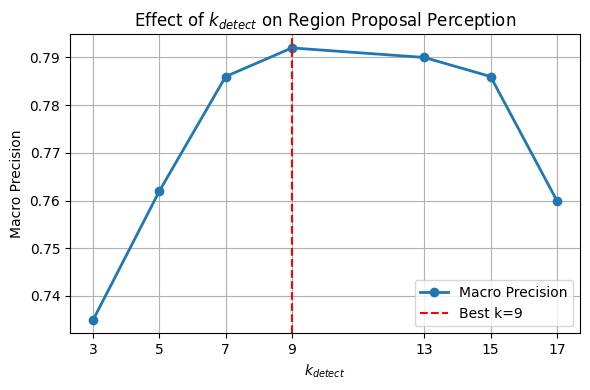}
    \caption{Effect of $k_{\text{detect}}$ on macro precision (post self-refinement). The peak performance is achieved at $k=9$, validating its use as the default setting for region proposal evaluation.}
    \label{fig:k_detect_plot}
\end{figure}

We evaluated macro precision on the PASCAL VOC dataset using region proposals generated by a pretrained Faster R-CNN detector, with $k_{\text{detect}}$ values ranging from 3 to 17. The results are presented in Figure~\ref{fig:k_detect_plot}. The model achieves the highest macro precision of 0.792 at $k=9$. Smaller values of $k$ underperform due to insufficient neighborhood support for robust label inference, while larger $k$ values introduce noisy or semantically irrelevant embeddings that dilute label confidence.

We therefore adopt $k_{\text{detect}} = 9$ as the default configuration in all cross-dataset experiments involving region proposal classification. This setting ensures strong performance while maintaining computational efficiency and neighborhood specificity.

\section{Qualitative Examples of Predicted Labels}
\label{sec:qualitative_examples}

\begin{figure}[h]
    \centering
    \includegraphics[width=\linewidth]{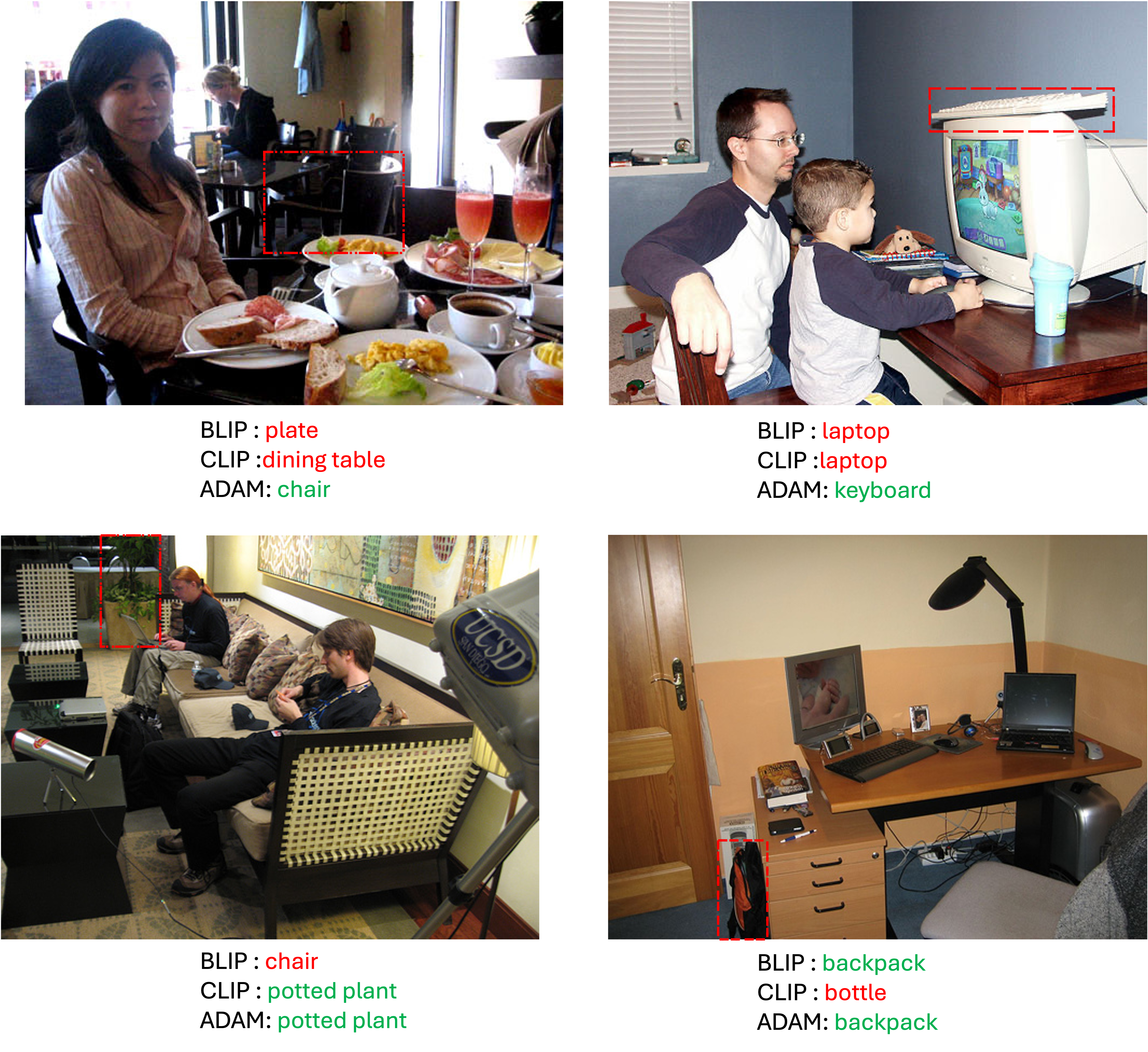}
    \caption{Qualitative examples of ADAM’s predictions on region proposals from PASCAL VOC. \textcolor{green}{Green text}: correct predictions. \textcolor{red}{Red text}: incorrect predictions.}
    \label{fig:qual_results}
\end{figure}

To illustrate the effectiveness of ADAM’s label prediction process in open-world scenarios, we present qualitative examples in Figure~\ref{fig:qual_results}. These examples are drawn from scenes in the PASCAL VOC dataset, where region proposals were generated using Faster R-CNN. ADAM performs zero-shot label assignment for these proposals using only visual similarity and contextual reasoning—without requiring any training or fine-tuning.

These results highlight ADAM's robustness in challenging scenarios such as occlusion, low resolution, and semantic overlap. Even when predictions deviate from ground truth, they often remain semantically plausible.

% \begin{figure}[t]
% \centering
% \includegraphics[width=\linewidth]{Precision-full.png} % Replace with your file name
% \caption{\textbf{Precision Comparison Between CLIP and ADAM Across VOC Classes.} ADAM (blue) outperforms CLIP (orange) in most categories when applied on Faster R-CNN region proposals, achieving a higher macro precision of 0.7925 versus 0.7487.}

% \label{fig:faster_prec}
% \end{figure}

%%%%%%%%%%%%%%%%%%%%%%%%%%%%%%%%%%%%%%%%%%%%%%%%%%%%%%%%%%%%